\documentclass[sn-basic]{sn-jnl}
 \pdfoutput=1

\jyear{2021}%

\theoremstyle{thmstyleone}%
\theoremstyle{thmstyletwo}%

\theoremstyle{thmstylethree}%

\usepackage{amsmath}
\usepackage{amsfonts} 
\usepackage{amssymb}
\usepackage{placeins}
\usepackage{graphicx}
\usepackage{float}
\usepackage{subfig}
\usepackage{lscape}
\usepackage{float}
\usepackage{hyperref}
\usepackage{booktabs}  
\usepackage{listings}

\usepackage{url}
\usepackage{natbib}
\usepackage[algo2e, ruled,vlined]{algorithm2e}

\definecolor{cornellred}{rgb}{0.7, 0.11, 0.11}
\definecolor{steelblue}{rgb}{0.2745, 0.5098, 0.7059}

\begin{document}

\title[Extended 
Missing Data Imputation via GANs for Ranking Applications]{Extended 
Missing Data Imputation via GANs for Ranking Applications}


\author*[1]{\fnm{Grace} \sur{Deng}}\email{gd345@cornell.edu}

\author[2]{\fnm{Cuize} \sur{Han}}\email{cuize@amazon.com}

\author[1]{\fnm{David S.} \sur{Matteson}}\email{dm484@cornell.edu}

\affil[1]{\orgdiv{Department of Statistics and Data Science}, \orgname{Cornell University}, \orgaddress{ \city{Ithaca}, \state{New York}, \country{USA}}}

\affil[2]{\orgdiv{Amazon Search},  \orgaddress{\city{Palo Alto}, \state{CA}, \country{USA}}}

%\affil[3]{\orgdiv{Department}, \orgname{Organization}, \orgaddress{\street{Street}, \city{City}, \postcode{610101}, \state{State}, \country{Country}}}


\abstract{
We propose Conditional Imputation GAN, an extended missing data imputation method based on Generative Adversarial Networks (GANs). The motivating use case is learning-to-rank, the cornerstone of modern search, recommendation system, and information retrieval applications. Empirical ranking datasets do not always follow standard Gaussian distributions or Missing Completely At Random (MCAR) mechanism, which are standard assumptions of classic missing data imputation methods. Our methodology provides a simple solution that offers compatible imputation guarantees while relaxing assumptions for missing mechanisms and sidesteps approximating intractable distributions to improve imputation quality. We prove that the optimal GAN imputation is achieved for Extended Missing At Random (EMAR) and Extended Always Missing At Random (EAMAR) mechanisms, beyond the naive MCAR. Our method demonstrates the highest imputation quality on the open-source Microsoft Research Ranking (MSR) Dataset and a synthetic ranking dataset compared to state-of-the-art benchmarks and across various feature distributions. Using a proprietary Amazon Search ranking dataset, we also demonstrate comparable ranking quality metrics for ranking models trained on GAN-imputed data compared to ground-truth data. 

%We propose a missing data imputation method based on Conditional Generative Adversarial Networks (GANs) for a novel e-commerce application: learning-to-rank with incomplete training data. Classical imputation methods often have strict assumptions on underlying missing mechanisms and feature distributions. Our methodology provides a simple solution that offers compatible imputation guarantees across different missing mechanisms, sidesteps approximating intractable distributions to improve imputation quality, and supports downstream business applications. We prove that the optimal GAN imputation offers theoretical guarantees for missing mechanisms Extended Missing At Random (EMAR) and Extended Always Missing At Random (EAMAR), beyond the naive Missing Completely At Random (MCAR). Using a proprietary Amazon Search ranking dataset, GAN imputations produce the lowest RMSE across all levels of missingness. Using GAN-completed ranking dataset, we train production-level ranking models that are comparable to training on ground-truth data based on ranking quality metrics NDCG and MRR. Finally, we replicate this success using the open-source Microsoft Research Ranking (MSR) Dataset and a synthetic ranking dataset, with superior performance against non-GAN benchmarks and across various types of feature distributions.
}




\maketitle

\section{Introduction}
Missing data is a prevalent data quality issue found in all aspects of data science and machine learning. Modern data collection technology can often exhibit non-random gaps in data due to a variety of reasons, e.g., non-response bias. At the same time, many machine learning models require complete datasets for training, highlighting the need for missing data imputation methods that are broadly applicable to different types of datasets characterized by complex missing mechanisms.

\subsection{Motivation}
Our motivating application is the classic ``learning-to-rank" problem for search, recommendation systems, and information retrieval \citep{li2011short, burges2010ranknet}. The ranking dataset has a unique structure compared to panel, time series, or image datasets. It is characterized by query-groups, where individual results are associated with a query and ordered by a ranking model, and composite features with non-standard distributions that will vary both on the query-group and query-result level. Training on missing data leads to biased ranking models \citep{marlin2009collaborative} and dropping individual query-results with missing values is difficult given the ordered nature of ranking data.

\begin{figure}[htp]
    \centering
    \includegraphics[width=0.9\textwidth]{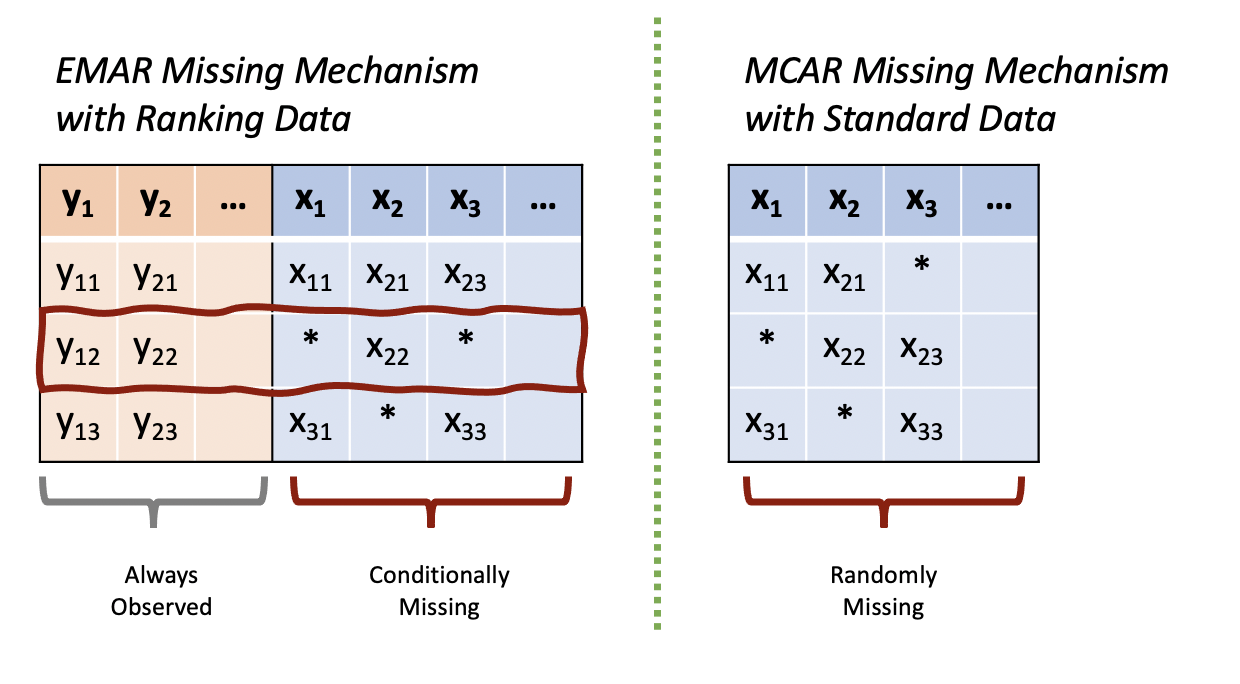}
    \caption{Missing data scenarios under EMAR vs MCAR mechanisms. Empirical ranking dataset with always observed columns is example of EMAR; see Section \ref{sec:emar_mar} for formal definitions.}  
    \label{fig:emar_vs_mcar}
\end{figure}

A further challenge for imputation in ranking applications is violation of the Missing Completely At Random (MCAR) \citep{heitjan1996distinguishing, doretti2018missing} and Gaussian distribution assumptions favored by classic imputation methods; see Figure \ref{fig:emar_vs_mcar}. Ranking datasets can include columns that are always observed, which influences the probability of missingness of other features, e.g., a product labeled ``New" has more missing feature values due to lack of data. Meanwhile, the MCAR mechanism requires the probability of missingness to be independent of all data values. A more appropriate mechanism for ranking datasets would be Missing At Random (MAR), which is less restrictive and specifies that missingness only depends on observed components \citep{little2019statistical}. Finally, the mechanism is called Missing Not At Random (MNAR) if missingness depends on unobserved components. 
%We formally define these and extended missing mechanisms in Sections \ref{sec:def_mar_amar} and \ref{sec:emar_mar}. 
%\textcolor{red}{Explain limitations of MICE and MissForest, and cite stuff for more recent methods.}

Currently, standard imputation methods such as MICE and MissForest have strict assumptions on underlying missing mechanisms and feature distributions \citep{buuren2010mice, stekhoven2012missforest}. Alternatively, using prediction methods for missing values requires a custom model per feature as well as a set of predictors that are never missing; this is near impossible to achieve with real data. More recent methods \citep{yoon2018gain, li2019misgan, luo2018multivariate}, involving generative models do not necessarily address more complex dataset structures or account for auxiliary information that influence the underlying data generating process.
%the unique nature of complex dataset structures (e.g., ranking) and heterogeneity in feature distributions conditional on auxiliary information.

We propose a novel extended missing data imputation method by adapting Conditional Generative Adversarial Networks (CGANs) \citep{goodfellow2014generative, mirza2014conditional, yoon2018gain}. Our method aims to encompass more complex data structures and missing mechanisms; empirical ranking datasets are a prime example given its ``ordered-grouping" characteristics and heterogeneous distributions across different query-groups. Furthermore, we define more realistic missing scenarios Extended Missing At Random (EMAR) and Extended Always Missing At Random (EAMAR) based on MAR and show that GAN-generated imputations satisfy conditions for compatible imputations under these new mechanisms.

\subsection{Related Methods}
GAN architecture is comprised of two competing deep neural nets; the \textit{generator} and the \textit{discriminator}. The generator produces synthetic data by mimicking the underlying data distribution, and the discriminator tries to distinguish fake and real data. Training GANs is a balancing act in that neither system should dominate the other too quickly. A strong generator may lead to mode collapse \citep{thanh2020catastrophic} by memorizing select samples, while a strong discriminator can lead to near-zero gradients and non-convergence by perfectly classifying samples.

GANs have shown impressive performance in generative tasks, including high-resolution image generation, text-to-image synthesis, image-to-image translation, video synthesis, and audio generation/synthesis \citep{oza2019progressive, sheng2019unsupervised}. A `vanilla' GAN has a tendency to suffer from non-convergence and mode collapse. The introduction of Deep-Convolutional GANs (DC-GAN) with convolution layers \citep{radford2015unsupervised} greatly improved the stability of GANs during training and showed that the generator and discriminator learned a hierarchy of representations. Other techniques for GAN training include feature mapping, batch normalization \citep{salimans2016improved}, leaky-relu activation functions, and modifications of the objective loss functions, e.g., Wasserstein loss \citep{arjovsky2017wasserstein}. Conditional GANs \citep{mirza2014conditional} allow for greater control over the modes of generated data by conditioning on auxiliary information such as class labels. 

 The Generative Adversarial Imputation Networks (GAIN) algorithm was first proposed \citep{yoon2018gain} to address this problem under the naive MCAR %(Missing Completely At Random) 
 assumption. This algorithm generated better imputations benchmarked against distributive missing data methods such as MICE \citep{buuren2010mice} and MissForest \citep{stekhoven2012missforest}, and generative methods such as Expectation-Maximization (EM). These methods are limited by assuming an underlying parametric distribution for missing data \citep{van2018flexible}, and MICE in particular assumes MCAR. Imputation with GANs addresses the gaps in current imputation methods where the generator strives to accurately impute missing data, and the discriminator strives to distinguish between observed and imputed data while minimizing the traditional minimax loss function. Applied to ranking, there are two drawbacks: the restrictive assumption of MCAR missing mechanism and the inability to account for heterogeneous subgroups with different data distributions.
 
Other methods are limited to images. Mis-GAN \citep{li2019misgan} utilized two separate GANs for data and mask (a matrix for indicating missing values) imputation for images, under various data corruption scenarios. Colla-GAN \citep{lee2019collagan} proposed converting the image imputation problem to a multi-domain images-to-image translation problem, resulting in imputations with higher visual quality. GAMIN \citep{yoon2020gamin} specifically targets high missingness levels ($>85\%$). Non-image data use cases include imputation for sequential data such as multivariate time series \citep{luo2018multivariate, kim2020survey, guo2019data, zhang2021missing}.

\subsection{Our Approach \& Contributions}
We propose an extended Conditional Imputation GAN with three key contributions:
\begin{enumerate}
    \item We introduce two new missing mechanisms, EMAR and EAMAR, that encompass broader empirical dataset types and provide theoretical guarantees for compatible imputations via GANs.
    \item We propose a Conditional Imputation GAN that allows for flexible imputation across (i) different data distributions, (ii) heterogeneous subgroups based on auxiliary information, and (iii) our new extended missing mechanisms. 
    \item We illustrate the superior imputation quality of our method against state-of-the-art benchmarks using open-source Microsoft Research ranking dataset and a proprietary 1.8 million query-group Amazon Search dataset. 
\end{enumerate}

To our knowledge, there has been no prior work exploring GAN imputation for machine-learned ranking (MLR) applications. Our method greatly expands the theoretical basis for GAN-based imputation methods for complex datasets and missing mechanisms, and is the first to adapt Conditional GANs for imputation on industry-scale ranking datasets. 

%address the learn-to-rank problem with missing data through Conditional GANs for imputation, and to show that ranking models trained on imputed data is empirically comparable to ground-truth models. 

Through empirical evaluations, we showcase the superior imputation quality of our method against benchmarks using three ranking datasets: a public Microsoft Research\footnote{Data available at https://www.microsoft.com/en-us/research/project/mslr/} ranking dataset with heterogeneous subgroups, a simulated ranking dataset with extensive feature distributions, and a proprietary 1.8 million query-group Amazon Search dataset. The Conditional Imputation GAN is particularly effective for imputing data under non-MCAR scenarios with non-standard distributions, as well as being computationally efficient for large-scale datasets.

As an investigation to downstream application impact, we also train standard ranking models on the imputed data versus the ground-truth data based on a shared target (e.g., clicks or purchases). We then evaluate standard ranking quality measures such as Normalized Discounted Cumulative Gain (NDCG) and Mean Reciprocal Rank (MRR). Our results demonstrate standard ranking models trained on imputed data has comparable performance to models trained on ground-truth complete data, indicating potential broader applicability to other business applications that are also impacted by pervasive data quality (missingness) problems.

\section{Methodology}
We briefly summarize Conditional GANs and how they can be adapted for imputation that better reflects non-MCAR missingness and heterogeneous feature distributions in ranking datasets. New missing mechanisms EMAR and EAMAR are introduced, and theoretical analysis provided for compatible imputations via the Conditional Imputation GAN.

\subsection{Conditional GAN}
Standard GANs consist of two adversarial models: a generator $G$ that mimics the true data distribution $p_{data}$ and a discriminative model $D$ that predicts the probability that a sample comes from the true distribution or the generated distribution $p_G$ from $G$ \citep{goodfellow2014generative}. The models $G$ and $D$ can theoretically be any non-linear mapping function, such as deep neural nets, with a variety of tuning parameters and configurations.
%During training, the generator $G$ maps a noise distribution $p_{Z}(Z)$ to the data space $G(Z; \theta_{G})$ and the discriminator $D$ predicts the probability that the data $X$ comes from training data. 
This is set-up as a two-player min-max game with value function $V(D,G)$:
\begin{equation}
\begin{split}
     \min_{G}\max_{D}V(D,G) &= \mathbb{E}_{X \sim p_{data}(X)}[\log D(X)] \\
     &+ \mathbb{E}_{Z\sim p_{Z}(Z)}[\log(1-D(G(Z)))]
\end{split}
\end{equation}

Suppose that there is auxiliary information $Y$ about the data $X$. We modify the standard GAN structure by conditioning on $Y$ in both the discriminator and generator, and combine the input noise $p_{Z}(Z)$ and $Y$ as a joint hidden representation. The resulting Conditional GAN (CGAN) \citep{mirza2014conditional} value function then becomes:

\begin{equation}
\begin{split}
    \min_{G}\max_{D}V(D,G) &= \mathop{\mathbb{E}_{X \sim p_{data}(X)}}\log(D(X \vert Y)) 
    \\ &+ \mathbb{E}_{Z\sim p_{Z}(Z)}(\log(1-D(G(Z \vert Y))))
\end{split}
\end{equation}

\subsection{Conditional Imputation GAN }
We introduce the Conditional Imputation GAN structure after briefly summarizing the GAIN structure\footnote{Code available at https://github.com/jsyoon0823/GAIN} \citep{yoon2018gain}. First we define $\boldsymbol{X} = (X_1, ..., X_d)$ as a random vector that could take on either continuous or discrete values (ranking features), and our training data are realizations of $\boldsymbol{X}$. We define the random vector $\boldsymbol{M}$ with the same dimensions as $\boldsymbol{X}$ which takes on values in $\{0, 1\}^{d}$; this is the missingness indicator matrix \citep{little2019statistical}, or mask matrix for short. Define a new random vector for observed data $\tilde{\boldsymbol{X}}$ as follows:
\begin{equation}
    \tilde{X}_{i} = \begin{cases}
X_{i}, \quad \text{if $M_{i}$ = 1}\\
*, \quad   \quad \text{otherwise}\\
\end{cases}
\end{equation}{}
where $*$ represents an unobserved or missing value replaced by noise value $Z_{i}$. Hence, $M$ explicitly indicates which values of $\tilde{X}_{i}$ are observed ($M_{i}$ = 1) and which are missing ($M_{i}$ = 0). 

$\tilde{\boldsymbol{X}}$, $\boldsymbol{M}$, and $\boldsymbol{Z}$ are now inputs into the generator $G$, which will generate an output vector of imputations $\bar{\boldsymbol{X}} = G(\tilde{\boldsymbol{X}}, \boldsymbol{M}, (\boldsymbol{1}-\boldsymbol{M}) \odot \boldsymbol{Z})$ of the same dimension as $\tilde{\boldsymbol{X}}$. The function $\odot$ denotes element-wise multiplication. Note that $\boldsymbol{Z}$ is independent of all other variables; it can be Gaussian noise but can also be designated otherwise depending on the dataset. $\bar{\boldsymbol{X}}$ is the imputed copy of $\tilde{\boldsymbol{X}}$, but we are only interested in the values of $\bar{\boldsymbol{X}}_{i}$ for which ${M}_{i} = 0$, that is, when the value is unobserved. Hence, the completed data vector $\hat{\boldsymbol{X}}$ is
\begin{equation}
\begin{split}
    \hat{\boldsymbol{X}} &= \boldsymbol{M} \odot \tilde{\boldsymbol{X}}  
    + (1-\boldsymbol{M}) \odot \bar{\boldsymbol{X}}
\end{split}
\end{equation}
The discriminator $D$ then tries to recover the true $\boldsymbol{M}$ from the completed data vector $\hat{\boldsymbol{X}}$, by predicting the probability of whether each $\hat{\boldsymbol{X}}_{i}$ is real (observed) or fake (imputed). The resulting vector of probabilities is denoted as $\hat{\boldsymbol{M}} = D(\hat{\boldsymbol{X}})$. Hence, the goal of the generator-discriminator pair is to minimize the distance between $\boldsymbol{M}$ and $\hat{\boldsymbol{M}}$. 

Given an arbitrary loss function $\mathcal{L}$, the value function is a two-player min-max game. Using the cross-entropy loss function gives:

\begin{equation}
\begin{split}
    \min_{G}\max_{D}V(D,G) &= \mathbb{E}[\mathcal{L}(\boldsymbol{M}, \hat{\boldsymbol{M}})] \\ 
    &= \mathbb{E}[\sum^{n}_{i=1} (\boldsymbol{M}_{i}\log(\hat{\boldsymbol{M}}_{i}) \\ &+ (1-\boldsymbol{M}_{i})\log(1-\hat{\boldsymbol{M}}_{i}))]
\end{split}
\end{equation}

However, this set-up is too naive and fails to account for heterogeneity across different subsets or class labels within many real-world datasets; hence, we propose adapting Conditional GANs to address these concerns.

\begin{figure}[htp]
    \centering
    \includegraphics[width=\textwidth]{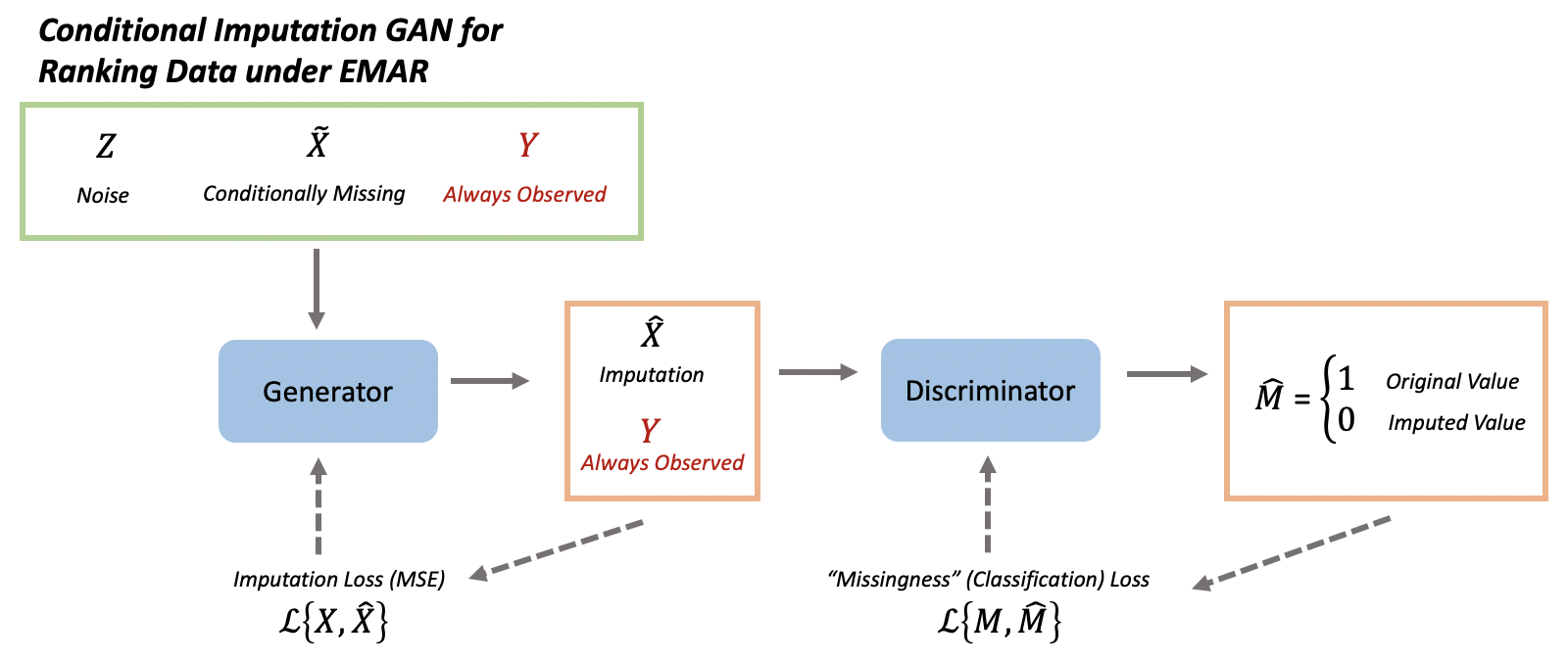}
    \caption{Conditional Imputation GAN Overview.}
    \label{fig:gan_flowchart}
\end{figure}

Suppose we have auxiliary information $\boldsymbol{Y}$ that is always observed along with data $\tilde{\boldsymbol{X}}$ that is conditionally missing under EMAR; $\boldsymbol{Y}$ may influence the probability of missingness or underlying data distribution in other features. We then condition on $\boldsymbol{Y}$ by feeding $\tilde{\boldsymbol{X}}$, $\boldsymbol{M}$, $\boldsymbol{Z}$, and $\boldsymbol{Y}$ into the generator $G$. This will generate an output vector of imputations $ \bar{\boldsymbol{X}}$ = $G(\tilde{\boldsymbol{X}} , \boldsymbol{M}, (1-\boldsymbol{M}) \odot \boldsymbol{Z}\vert \boldsymbol{Y}) $, which we then use to form the completed data vector $\hat{\boldsymbol{X}}$ in (4) and feed into the discriminator in order to recover $\boldsymbol{M}$. The output of probabilities is now $\hat{\boldsymbol{M}} = D(\hat{\boldsymbol{X}}\vert \boldsymbol{Y})$. Finally, we write the objective function of the Conditional Imputation GAN as: 

\begin{equation}
 \min_{G}\max_{D}V(D,G) = \mathbb{E}[\mathcal{L}(\boldsymbol{M}, \hat{\boldsymbol{M}})] = \mathbb{E}[\mathcal{L}(\boldsymbol{M},D(\hat{\boldsymbol{X}}\vert \boldsymbol{Y}))]
\label{eq: cond_gan_loss_func}
\end{equation} 
%\vspace{-2mm}
Using \ref{eq: cond_gan_loss_func}, we expand GAN imputation for empirical ranking datasets by conditioning on columns that are always observed during training and separating them from the imputation loss function. See Figure \ref{fig:gan_flowchart} and Algorithm \ref{algorithm: psuedo_code} for details.

\begin{algorithm}[h]
\SetAlgoLined
%\KwResult{stuff}
Given mini-batch size $n_{mb}$ \\ \;
\For{number of training iterations}{
Draw $n_{mb}$ samples $(\tilde{\boldsymbol{x}} , \boldsymbol{m}, \boldsymbol{z}, \boldsymbol{y})$ from $\tilde{\boldsymbol{X}}, \boldsymbol{M}, \boldsymbol{Z}$, and $\boldsymbol{Y}$ \\ \;
\textbf{(1) Generator $G$} \\ \;
Generate imputation $\bar{\boldsymbol{x}} \leftarrow G(\tilde{\boldsymbol{x}} , \boldsymbol{m}, \boldsymbol{z}, \boldsymbol{y})$  \\ \;
Complete data $\hat{\boldsymbol{x}} \leftarrow \boldsymbol{m} \odot \tilde{\boldsymbol{x}} + (1-\boldsymbol{m}) \odot \bar{\boldsymbol{x}}$ \\ \;
Compute $\mathcal{L}_{G} = -\sum(1-\boldsymbol{m}) \odot \log(\hat{\boldsymbol{m}})$ where $\hat{\boldsymbol{m}} = D(\hat{\boldsymbol{x}}, \boldsymbol{y})$ \\ \;
Update $G$ using Adam optimizer \;
 
\textbf{(2) Discriminator $D$} \\ \;
Predict $\hat{\boldsymbol{m}} \leftarrow D(\hat{\boldsymbol{x}}, \boldsymbol{y})$  \\  \;
Compute $\mathcal{L}_{D} = -\sum [\boldsymbol{m} \odot \log(\hat{\boldsymbol{m}}) + (1-\boldsymbol{m}) \odot \log(1-\hat{\boldsymbol{m}})]$ \\ \;
Update $G$ using Adam optimizer\;}
\caption{Psuedo-code for Conditional Imputation GAN}
\label{algorithm: psuedo_code}
\end{algorithm}

\subsection{Theoretical Analysis}
Prior works trying to extend or improve GAN imputation \citep{lee2019collagan, li2019misgan, camino2019improving, kim2020survey} all restricted theoretical guarantees for the generated distributions to the Missing Completely At Random (MCAR) assumption. In practice, this is too restrictive and rarely satisfied by real-world missing data. Empirically, we often find GAN imputations working quite well for missing data under MAR or even MNAR, and there is a clear gap between theoretical guarantees and empirical results. Here, we aim to close this gap by investigating more general conditions on missing mechanisms and extend theories beyond the MCAR assumption. %We focus on the vanilla GAN for imputation (the GAIN algorithm) for simplicity, and all the analysis also apply to Conditional GAN for imputation. 

In the following analysis, we use small case letters to represent the
$n$ independent realizations of $\boldsymbol{X},\boldsymbol{M}$
and $\tilde{\boldsymbol{X}}$ as $\boldsymbol{x}_{i}=(x_{i1,},...,x_{id})$,
$\boldsymbol{m}_{i}=(m_{i1,},...,m_{id})$ and $\boldsymbol{\tilde{x}}_{i}=(\tilde{x}_{i1,},...,\tilde{x}_{id})$, $i=1,..,n$. For a vector $\boldsymbol{x}$ of dimension $d$, we use the notation $\boldsymbol{x}\vert_{\boldsymbol{m}}$ to
represent the subvector of $\boldsymbol{x}$ that corresponds to the positions where the elements of $\boldsymbol{m}$ is $1$, i.e., the observed data components. Curly brackets within conditional probability statements are used for readability.

First we state the main theoretical result in \cite{yoon2018gain} which we will utilize to extend the theoretical analysis. A necessary and sufficient condition for $\hat{\boldsymbol{X}}$ being generated by an ideal generator is
\begin{equation}
\mathbb{P}\!\left(\hat{\boldsymbol{X}}=\boldsymbol{x} \ \middle \vert \ \{\boldsymbol{H}=\boldsymbol{h},M_{i}=t \} \right )=\mathbb{P}\!\left(\hat{\boldsymbol{X}}=\boldsymbol{x} \ \middle \vert \ \boldsymbol{H}=\boldsymbol{h}  \right)
\label{eq:condition for ideal G}
\end{equation}
for every $i\in\{1,...,d\},t\in\{0,1\},\boldsymbol{x}\in\mathcal{X}^{d}$ and $h\in\mathcal{H}$ such that $\mathbb{P}(\boldsymbol{H}=\boldsymbol{h} \ \vert \ M_{i}=t)>0$. Here $\boldsymbol{H}$ is a hint mechanism that takes value in the space $\mathcal{H}$ and it is a random vector
defined by us given $\boldsymbol{M}$ and $\tilde{\boldsymbol{X}}$.
$\mathbb{P}$ is the underlying probability measure and to keep notations simple, we assume, without loss of generality, all the random variables involved are discrete. Note that this result holds without any assumptions on the joint distribution of $(\boldsymbol{X},\boldsymbol{M})$. Hence, we can utilize this to extend beyond the naive MCAR assumption.

\subsubsection{MAR and AMAR - Missing Mechanism} \label{sec:def_mar_amar}
A missing model is the specification of the conditional distribution
$\boldsymbol{M} \vert \boldsymbol{X}$ which governs the missing data generation
process. A missing mechanism is certain assumptions made to $\boldsymbol{M} \vert \boldsymbol{X}$ that can be satisfied by a set of missing models. Three classic types of missing mechanisms are MCAR, MAR (Missing At Random) and MNAR (Missing Not At Random) \citep{little2019statistical}. Roughly speaking, MCAR means whether the data is missing or not is independent of the data, MAR requires the probability of missingness only depends on the observed data and MNAR allows missingness to depend on unobserved data. There are some subtleties in the definition of MAR. As it involves the observed data, do we mean that the assumption is only being made on our realized sample at hand $(\boldsymbol{\tilde{x}}_{i},\boldsymbol{m}_{i}),i=1,.,n$ or on any future sample that we may observe? Clearly the latter is a stronger assumption. This has been made clear and discussed thoroughly in \cite{seaman2013meant} and  \cite{mealli2015clarifying}. We follow \cite{mealli2015clarifying} to define MAR as assuming probability of missingness depends only on realized samples, and AMAR (Always Missing At Random) as depending on any future sample. Formally, we say $(\boldsymbol{X},\boldsymbol{M})$ is \textbf{MAR} given the
realized sample $(\boldsymbol{\tilde{x}}_{i},\boldsymbol{m}_{i}),i=1,...,n$ if 
\begin{equation}
\mathbb{P}\! \left(\boldsymbol{M}=\boldsymbol{m}_{i} \ \middle \vert \ \{ \boldsymbol{X}\vert_{\boldsymbol{m}_{i}}=\boldsymbol{\tilde{x}}_{i}\vert_{\boldsymbol{m}_{i}} \} \right)=\mathbb{P}\! \left(\boldsymbol{M}=\boldsymbol{m}_{i} \ \middle\vert \ \boldsymbol{X}=\boldsymbol{x} \right)
\label{eq: MAR1}
\end{equation}
for any $i=1,..,n$ and $\boldsymbol{x}$ such that $\boldsymbol{x}\vert_{\boldsymbol{m}_{i}}=\boldsymbol{\tilde{x}}_{i}\vert_{\boldsymbol{m}_{i}}$.

$(\boldsymbol{X},\boldsymbol{M})$ is \textbf{AMAR} if
\begin{equation}
\mathbb{P}(\boldsymbol{M}=\boldsymbol{m} \  \vert  \ \{ \boldsymbol{X}\vert_{\boldsymbol{m}}=\boldsymbol{x}\vert_{\boldsymbol{m}}\})=\mathbb{P}(\boldsymbol{M}=\boldsymbol{m} \ \vert \ \boldsymbol{X}=\boldsymbol{x})\label{eq: AMAR1}
\end{equation}
for any $\boldsymbol{m}\in\{0,1\}^{d}$ and $\boldsymbol{x}\in\mathcal{X}^{d}$. By the property of conditional probability and the fact that $\{\boldsymbol{X}=\boldsymbol{x}\}$ implies $\{\boldsymbol{X}\vert_{\boldsymbol{m}}=\boldsymbol{x}\vert_{\boldsymbol{m}}\}$, (\ref{eq: MAR1}) and (\ref{eq: AMAR1}) are equivalent to 
\begin{equation}
\begin{split}
    \mathbb{P}(\boldsymbol{X}=\boldsymbol{x} \ \vert \ \{ \boldsymbol{X}\vert_{\boldsymbol{m}_{i}}=\boldsymbol{\tilde{x}}_{i}\vert_{\boldsymbol{m}_{i}},\boldsymbol{M}=\boldsymbol{m}_{i} \}) \\ =\mathbb{P}(\boldsymbol{X}=\boldsymbol{x} \ \vert\ \{\boldsymbol{X}\vert_{\boldsymbol{m}_{i}}=\boldsymbol{\tilde{x}}_{i}\vert_{\boldsymbol{m}_{i}}\})
\end{split}
\label{eq: eqv MAR}
\end{equation}
 and 
\begin{equation}
\begin{split}
    \mathbb{P}(\boldsymbol{X}=\boldsymbol{x}\ \vert\ \{ \boldsymbol{X}\vert_{\boldsymbol{m}}=\boldsymbol{x}\vert_{\boldsymbol{m}},\boldsymbol{M}=\boldsymbol{m} \}) \\ =\mathbb{P}(\boldsymbol{X}=\boldsymbol{x} \ \vert \ \{\boldsymbol{X}\vert_{\boldsymbol{m}}=\boldsymbol{x}\vert_{\boldsymbol{m}}\})
\end{split}
\label{eq: eqv AMAR}
\end{equation}
Given a fixed $\boldsymbol{m}_{0}\in\{0,1\}^{d}$, we emphasize here
that both conditions do not imply the conditional independence of $\boldsymbol{X}$
and $\boldsymbol{M}$ given $\boldsymbol{X}\vert_{\boldsymbol{m}_{0}}$
which is a much stronger assumption that requires 
\begin{equation}
\begin{split}
    \mathbb{P}(\boldsymbol{X}=\boldsymbol{x} \ \vert \ \{ \boldsymbol{X}\vert_{\boldsymbol{m}_{0}}=\boldsymbol{x}\vert_{\boldsymbol{m}_{0}},\boldsymbol{M}=\boldsymbol{m} \} ) \\ =\mathbb{P}(\boldsymbol{X}=\boldsymbol{x} \ \vert \ \{ \boldsymbol{X}\vert_{\boldsymbol{m}_{0}}=\boldsymbol{x}\vert_{\boldsymbol{m}_{0}} \})
\end{split}
\label{eq: conditional independence condition}
\end{equation}
for any $\boldsymbol{m}\in\{0,1\}^{d}$ and $\boldsymbol{x}\in\mathcal{X}^{d}$. Many different missing mechanisms can be defined through those conditional
probability equations where different mechanisms correspond to different restrictions on the set of variable values that satisfy the equations. See \cite{doretti2018missing} for more examples.

\subsubsection{Compatible Imputations}
Prior work \citep{yoon2018gain} showed that, under the MCAR assumption, the ideal imputation $\hat{\boldsymbol{X}}$ has the same distribution as the original data. This is perfect but may be too stringent if we only care about the imputation quality for the missing data given the observed data.  Thus, we define two compatible conditions for imputation. We say $\hat{\boldsymbol{X}}$ is a \textbf{compatible} imputation for the missing data $(\boldsymbol{\tilde{x}}_{i},\boldsymbol{m}_{i}),i=1,...,n$
if 
\begin{equation}
\begin{split}
\mathbb{P}(\hat{\boldsymbol{X}}=\boldsymbol{x} \ \vert \ \{\boldsymbol{X}\vert_{\boldsymbol{m}_{i}}=\boldsymbol{\tilde{x}}_{i}\vert_{\boldsymbol{m}_{i}},\boldsymbol{M}=\boldsymbol{m}_{i} \} )= \\
 \mathbb{P}(\boldsymbol{X}=\boldsymbol{x} \ \vert \ \{\boldsymbol{X}\vert_{\boldsymbol{m}_{i}}=\boldsymbol{\tilde{x}}_{i}\vert_{\boldsymbol{m}_{i}},\boldsymbol{M}=\boldsymbol{m}_{i} \} )
 \end{split} 
 \label{eq: compatible for missing data}
\end{equation}
for any $i=1,..,n$ and $\boldsymbol{x}$ such that $\boldsymbol{x}\vert_{\boldsymbol{m}_{i}}=\boldsymbol{\tilde{x}}_{i}\vert_{\boldsymbol{m}_{i}}$.
We say $\hat{\boldsymbol{X}}$ is \textbf{always compatible} for imputing $(\boldsymbol{X},\boldsymbol{M})$
if 
\begin{equation}
\begin{split}
\mathbb{P}(\hat{\boldsymbol{X}}=\boldsymbol{x} \ \vert \ \{\boldsymbol{X}\vert_{\boldsymbol{m}}=\boldsymbol{x}\vert_{\boldsymbol{m}},\boldsymbol{M}=\boldsymbol{m} \} )= \\ 
\mathbb{P}(\boldsymbol{X}=\boldsymbol{x \ }\vert \ \{ \boldsymbol{X}\vert_{\boldsymbol{m}}=\boldsymbol{x}\vert_{\boldsymbol{m}},\boldsymbol{M}=\boldsymbol{m} \} )    
\end{split}
\label{eq: always compatible}
\end{equation}
for any $\boldsymbol{m}\in\{0,1\}^{d}$ and $\boldsymbol{x}\in\mathcal{X}^{d}$.
The compatible conditions are really what we desire for imputations. We will show that the GAN imputation $\hat{\boldsymbol{X}}$ still enjoys compatibility for many missing mechanisms beyond MCAR.

\subsubsection{EMAR and EAMAR - Extended Missing Mechanisms} \label{sec:emar_mar}
We formally define a new missing mechanism that we call \textbf{EMAR} (Extended Missing At Random). Formally, we say $(\boldsymbol{X},\boldsymbol{M})$ is
EMAR given the realized sample $(\boldsymbol{\tilde{x}}_{i},\boldsymbol{m}_{i}),i=1,...,n$
if 
\begin{equation} 
\begin{split}
    \mathbb{P}\left(\boldsymbol{X}=\boldsymbol{x} \ \middle\vert \ \{\boldsymbol{X}\vert_{\boldsymbol{m}_{i}}=\boldsymbol{\tilde{x}}_{i}\vert_{\boldsymbol{m}_{i}},\boldsymbol{M}=\boldsymbol{m} \} \right)   \\  =\mathbb{P}\left(\boldsymbol{X}=\boldsymbol{x} \ \middle\vert \ \{\boldsymbol{X}\vert_{\boldsymbol{m}_{i}}=\boldsymbol{\tilde{x}}_{i}\vert_{\boldsymbol{m}_{i}} \} \right)
\end{split}
\label{eq: EMAR}
\end{equation}
for any $i=1,..,n$, $\boldsymbol{x}$ such that $\boldsymbol{x}\vert_{\boldsymbol{m}_{i}}=\boldsymbol{\tilde{x}}_{i}\vert_{\boldsymbol{m}_{i}}$
and $\boldsymbol{m}=\boldsymbol{m}_{i}$ or $\boldsymbol{1}$. $(\boldsymbol{X},\boldsymbol{M})$ is \textbf{EAMAR} (Extended Always Missing At Random) if 
\begin{equation}
\begin{split}
    \mathbb{P}(\boldsymbol{X}=\boldsymbol{x} \ \vert \ \{\boldsymbol{X}\vert_{\boldsymbol{m}}=\boldsymbol{x}\vert_{\boldsymbol{m}},\boldsymbol{M}=\boldsymbol{m}' \} ) \\ =\mathbb{P}(\boldsymbol{X}=\boldsymbol{x} \ \vert \ \{\boldsymbol{X}\vert_{\boldsymbol{m}}=\boldsymbol{x}\vert_{\boldsymbol{m}}\} )
\end{split}\label{eq: EAMAR}
\end{equation}
for any $\boldsymbol{x}\in\mathcal{X}^{d}$ and $\boldsymbol{m}'=\boldsymbol{m}$ or $\boldsymbol{1}$. Examples of EMAR or EAMAR are much more prevalent in real datasets instead of MCAR, e.g., a ranking dataset with query-group columns that are always observed. For both missing mechanisms, we can then state: 

\textbf{Theorem 1} \ \emph{EMAR on $(\boldsymbol{X},\boldsymbol{M})$
given}\textbf{\emph{ }}\emph{the realized sample $(\boldsymbol{\tilde{x}}_{i},\boldsymbol{m}_{i}),i=1,...,n$
is a sufficient condition for $\hat{\boldsymbol{X}}$ being a compatible
imputation for the missing data $(\boldsymbol{\tilde{x}}_{i},\boldsymbol{m}_{i}),i=1,...,n$.
EAMAR on $(\boldsymbol{X},\boldsymbol{M})$ is a sufficient condition
for $\hat{\boldsymbol{X}}$ being always compatible for imputing $(\boldsymbol{X},\boldsymbol{M})$.}

\textbf{Proof}: Let $\boldsymbol{I}$ be a uniform random subset of $\{1,2,...,d\}$
that is independent with $(\boldsymbol{X},\boldsymbol{M},\hat{\boldsymbol{X}}).$
Given $\boldsymbol{I}$, define the random variable $J$ that uniformly
takes value in $\{1,2,...,d\}\backslash\boldsymbol{I}$. Given $\tilde{\boldsymbol{X}},\boldsymbol{M},\boldsymbol{I},J$, we define a pair of hint vectors $(\boldsymbol{H},\tilde{\boldsymbol{H}})$
: 
\begin{equation}
H_{j}=\begin{cases}
\tilde{X}_{j} & j\in\boldsymbol{I}\\
\# & j\notin\boldsymbol{I}
\end{cases}, \quad 
\tilde{H}_{j}=\begin{cases}
M_{j} & j\neq J\\
0.5 & j=J
\end{cases}
\end{equation}
where $\#\notin\mathcal{X}$ and it is different from the symbol $*$ that indicates missing. 

We will see that the proof for EAMAR implies always compatible and EMAR for realization implies compatible for realization is the same. We will focus on the former. Take any $\boldsymbol{x}\in{\mathcal X}^{d}$, $\boldsymbol{m}_{0}\in\{0,1\}^{d}$ and let $I_{0}=\{j:j\in\{1,2..,d\},m_{0j}=1\}$.
Also take $j_{0}\in\{1,2,...,d\}\backslash I_{0}.$ Let $\boldsymbol{h}=(h_{1},..,h_{d})$
and $\tilde{\boldsymbol{h}}=(\tilde{h}_{1},..,\tilde{h}_{d})$ to
be 
\[
h_{j}=\begin{cases}
\tilde{x}_{j} & j\in I_{0}\\
\# & j\notin I_{0}
\end{cases},
\quad \quad
\tilde{h}_{j}=\begin{cases}
m_{0j} & j\neq j_{0}\\
0.5 & j=j_{0}
\end{cases}
\]
From (\ref{eq:condition for ideal G}), we have 
\begin{equation}
\begin{aligned}
\mathbb{P}(\hat{\boldsymbol{X}}=\boldsymbol{x}  \vert \ \{\boldsymbol{H}=\boldsymbol{h},\tilde{\boldsymbol{H}}=\tilde{\boldsymbol{h}},M_{j_{0}}=0 \}) \\ = \mathbb{P}(\hat{\boldsymbol{X}}=\boldsymbol{x} \ \vert \ \{\boldsymbol{H}=\boldsymbol{h},\tilde{\boldsymbol{H}}=\tilde{\boldsymbol{h}},M_{j_{0}}=1 \}).
\end{aligned}
\label{eq: theorem2 1}
\end{equation}
Let $\boldsymbol{m}_{0}^{t}$ to be the vector that equals $\boldsymbol{m}_{0}$
component-wise except for $m_{0j_{0}}^{t}=t$. Note that the event
$\{\boldsymbol{H}=\boldsymbol{h},\tilde{\boldsymbol{H}}=\tilde{\boldsymbol{h}},M_{j_{0}}=t\}$
is equivalent to $\{\boldsymbol{I}=I_{0},J=j_{0},X_{j}=x_{j},\forall j\in I_{0},M_{j'}=m_{0j'},\forall j'\neq j_{0},M_{j_{0}}=t\}$. So we have 
\begin{equation}
\begin{aligned}
\mathbb{P}(\hat{\boldsymbol{X}}&=\boldsymbol{x} \ \vert\ \{\boldsymbol{H}=\boldsymbol{h},\tilde{\boldsymbol{H}}=\tilde{\boldsymbol{h}},M_{j_{0}}=t \}) \\ & =\mathbb{P}(\hat{\boldsymbol{X}}=\boldsymbol{x} \ \vert\ \{X_{j}=x_{0j},\forall j\in I_{0},M_{j'}=m_{0j'},\forall j'\neq j_{0},M_{j_{0}}=t \})\\
 & =\mathbb{P}(\hat{\boldsymbol{X}}=\boldsymbol{x}\ \vert \ \{\boldsymbol{X}\vert_{\boldsymbol{m}_{0}}=\boldsymbol{x}\vert_{\boldsymbol{m}_{0}},\boldsymbol{M}=\boldsymbol{m}_{0}^{t} \})
\end{aligned}
\label{eq: theorem2 2}
\end{equation}
where the first equality is because of $\text{\ensuremath{\boldsymbol{I}} }$ and $J$'s independence with other random variables. From (\ref{eq: theorem2 1})(\ref{eq: theorem2 2}), we have 
\begin{equation}
\begin{aligned}
\mathbb{P}(\hat{\boldsymbol{X}}=\boldsymbol{x} \ \vert \ \{\boldsymbol{X}\vert_{\boldsymbol{m}_{0}}=\boldsymbol{x}\vert_{\boldsymbol{m}_{0}},\boldsymbol{M}=\boldsymbol{m}_{0}^{1} \}) \\ 
=\mathbb{P}(\hat{\boldsymbol{X}}=\boldsymbol{x}\ \vert \ \{\boldsymbol{X}\vert_{\boldsymbol{m}_{0}}=\boldsymbol{x}\vert_{\boldsymbol{m}_{0}},\boldsymbol{M}=\boldsymbol{m}_{0}^{0} \})
\end{aligned}
\label{theorem2 3}
\end{equation}
By taking all the other $j_{0}'\in\{1,2,...,d\}\backslash I_{0}$ and follow the same procedure, we see that for any $\boldsymbol{m}\in\{0,1\}^{d}$
such that $\boldsymbol{m}\geq\boldsymbol{m}_{0}$ (componentwise), we have

\begin{equation}
\begin{aligned}\mathbb{P}(\hat{\boldsymbol{X}}&=\boldsymbol{x} \ \vert \ \{\boldsymbol{X}\vert_{\boldsymbol{m}_{0}}=\boldsymbol{x}\vert_{\boldsymbol{m}_{0}},\boldsymbol{M}=\boldsymbol{m}\}) \\
& =\mathbb{P}(\hat{\boldsymbol{X}}=\boldsymbol{x} \ \vert \ \{\boldsymbol{X}\vert_{\boldsymbol{m}_{0}}=\boldsymbol{x}\vert_{\boldsymbol{m}_{0}},\boldsymbol{M}=\boldsymbol{m}_{0}^{0} \})\\
 & =\mathbb{P}(\hat{\boldsymbol{X}}=\boldsymbol{x}\ \vert \ \{\boldsymbol{X}\vert_{\boldsymbol{m}_{0}}=\boldsymbol{x}\vert_{\boldsymbol{m}_{0}},\boldsymbol{M}=\boldsymbol{m}_{0}\})
\end{aligned}
\label{eq: theorem2 4}
\end{equation}
On the other hand, for the special case of $\boldsymbol{m}=\boldsymbol{1}$, we have 

\begin{equation}
\begin{aligned}\mathbb{P}(\hat{\boldsymbol{X}}&=\boldsymbol{x} \ \vert \ \{\boldsymbol{X}\vert_{\boldsymbol{m}_{0}}=\boldsymbol{x}\vert_{\boldsymbol{m}_{0}},\boldsymbol{M}=\boldsymbol{1}\}) \\
& =\mathbb{P}(\boldsymbol{X}=\boldsymbol{x} \ \vert \ \{\boldsymbol{X}\vert_{\boldsymbol{m}_{0}}=\boldsymbol{x}\vert_{\boldsymbol{m}_{0}},\boldsymbol{M}=\boldsymbol{1} \})\\
 & =\mathbb{P}(\boldsymbol{X}=\boldsymbol{x} \ \vert \ \{\boldsymbol{X}\vert_{\boldsymbol{m}_{0}}=\boldsymbol{x}\vert_{\boldsymbol{m}_{0}},\boldsymbol{M}=\boldsymbol{m}_{0}\})
\end{aligned}
\label{eq: theorem2 5}
\end{equation}
where the first equality holds because given $\boldsymbol{M}=\boldsymbol{1},$
$\hat{\boldsymbol{X}}=\boldsymbol{X}$ and the second one is due to
the EAMAR assumption (\ref{eq: EAMAR}). Thus combining (\ref{eq: theorem2 4})(\ref{eq: theorem2 5}), we prove the theorem.

To summarize, we demonstrated the advantages of Conditional Imputation GAN by showing the compatibility of optimal imputations under extended missing mechanisms EMAR and EAMAR. For the observed missing patterns, EMAR requires that the data distribution conditional on the observed values is the same as if they were not missing. EMAR is a stronger assumption compare to MAR, but it is much less restrictive than MCAR. In Theorem 1, we proved that EMAR, which includes a collection of missing models, is a sufficient condition for compatibility of optimal GAN imputation, and EAMAR is a sufficient condition for always compatibility. Whether the optimal GAN imputation is compatible under MNAR remains open for future work.

\section{Simulation - Imputation Quality by Data Distribution}
\subsection{Data and Methodology}
\begin{figure}[h]
    \centering
    \includegraphics[width=0.8\textwidth]{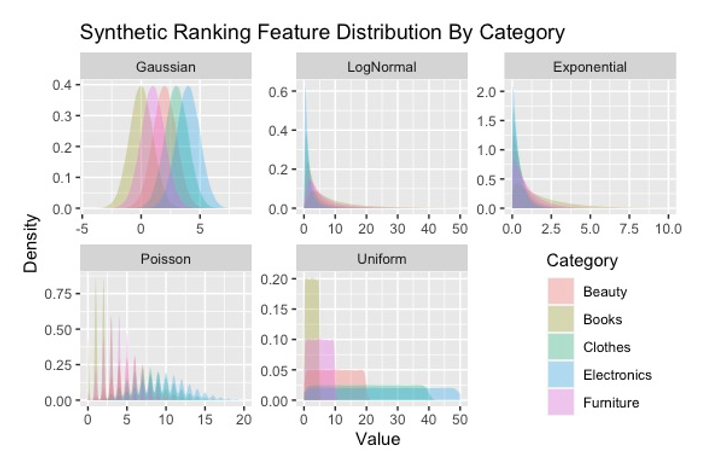}
    \caption{Synthetic - Ranking Feature Dist. by Category. Ranking features are simulated to follow both discrete and continuous distributions commonly found in empirical ranking datasets.}
    \label{fig:dist_synthetic}
\end{figure}

To illustrate how our method performs across a variety of data distributions, we first simulate a 10K query-group ranking dataset, with feature columns sampled from 5 distributions: Gaussian, LogNormal, Exponetnial, Poisson, Uniform. Each query-group has 64 query-results and is associated with a hypothetical product type (``Category") that is always observed, in accordance with the EMAR and EAMAR assumption. The 5 types are Books, Furniture, Beauty, Clothes, Electronics. The product category determines the true distribution parameters from which the ranking features are sampled; see Table \ref{table: synthetic_parameters} for details and Figure \ref{fig:dist_synthetic} for how ranking feature distributions vary by product type.  

\begin{table*}[htp] \centering 
  \caption{Synthetic Ranking Data: Feature Distribution by Category} 
  \label{table: synthetic_parameters} 
\resizebox{\columnwidth}{!}{
\begin{tabular}{@{\extracolsep{5pt}} cc|ccccc} 
\\[-1.8ex]\hline 
\hline \\[-1.8ex] 
 & & Books & Furniture & Beauty & Clothes & Electronics \\ 
\hline \\[-1.8ex] 
Gaussian & $N(\mu, \sigma^{2})$ & $N(0, 1)$ & $N(1, 1)$ & $N(2, 1)$ & $N(3, 1)$ & $N(4, 1)$ \\ 
LogNormal & $LN(\mu, \sigma^{2})$ & $LN(2, 1)$ & L$N(1.5, 1)$ & $LN(1, 1)$ & $LN(0.5, 1)$ & $LN(0, 1)$ \\ 
Exponential & $Exp(\lambda)$ & $Exp(0.5)$ & $Exp(1)$ & $Exp(1.5)$ & $Exp(2)$ & $Exp(2.5)$ \\ 
Poisson  & $Poisson(\lambda)$& $Pois(2)$ & $Pois(4)$ & $Pois(6)$ & $Pois(8)$ & $Pois(10)$ \\ 
Uniform & $Unif(a,b)$ & $Unif(0,2)$ & $Unif(0,4)$ & $Unif(0,6)$ & $Unif(0,8)$ & $Unif(0,10)$\\ 
\hline \\[-1.8ex] 
\end{tabular} 
}
\end{table*} 

%Describe methods and set-up, not results here.
We want to compare imputation quality as measured by RMSE across four methods: Conditional Imputation GAN, GAIN, MICE, and MissForest. Given that the Category column will always be observed, we select four levels of missingness (5\%, 10\%, 20\%, 30\%) and randomly mask feature values in each query-group as missing; this is aligned with the EMAR missing mechanism. 

For each method and missingness level, 10 imputations of the simulated ranking dataset are generated. In TensorFlow, each GAN replicate was trained for 50 epochs after normalizing the data and using the Adam optimizer. The generator and discriminator both utilized a standard architecture of fully-connected layers with leaky-relu activation. In R, default settings of MissForest and MICE are used, with only a 10\% random sample used for MissForest given the computational cost. We then compute average RMSE and standard errors over imputations across all ranking features and also separately (column-wise) for features from each distribution. See Table \ref{table: rmse_synthetic}.

\subsection{Simulation Results}

\begin{table}[htp] \centering 
  \caption{Simulation Results - Imputation Quality (RMSE) by Distributions} 
  \label{table: rmse_synthetic} 
   \resizebox{\columnwidth}{!}{
\begin{tabular}{@{\extracolsep{0pt}} cl|cccc} 
\\[-1.8ex]\hline 
\hline \\[-1.8ex] 
Distribution & Method & 5\% Missing & 10\% Missing & 15\% Missing & 20\% Missing  \\ 
\hline \\[-1.8ex] 
Overall & Cond. Imp. GAN & \textbf{0.841} ± (0.0002) & \textbf{0.861} ± (0.0001) & \textbf{0.893} ± (0.0001) & \textbf{0.91} ± (0.0001) \\ 
 & GAIN & 1.014 ± (0.0002) & 1.028 ± (0.0001) & 1.063 ± (0.0001) & 1.08 ± (0.0002) \\ 
 & MICE & 8.914 ± (0.1911) & 9.011 ± (0.1063) & 9.129 ± (0.2074) & 6.536 ± (0.0683) \\ 
 & MissForest & 6.536 ± (0.0683) & 6.514 ± (0.0831) & 7.018 ± (0.0338) & 7.018 ± (0.0338) \\ 
 \hline \\[-1.8ex] 
Exponential & Cond. Imp. GAN & \textbf{0.988} ± (0.0004) & \textbf{1.037} ± (0.0002) & \textbf{0.971} ± (0.0002) & \textbf{0.979} ± (0.0002) \\ 
 & GAIN & 1.161 ± (0.0003) & 1.044 ± (0.0002) & 1.009 ± (0.0001) & 1.004 ± (0.0001) \\ 
 & MICE & 1.584 ± (0.0313) & 1.574 ± (0.0144) & 1.587 ± (0.0113) & 1.068 ± (0.033) \\ 
 & MissForest & 1.068 ± (0.033) & 1.18 ± (0.0174) & 1.212 ± (0.0173) & 1.212 ± (0.0173) \\ 
Gaussian & Cond. Imp. GAN & \textbf{0.707} ± (0.0003) & \textbf{0.715} ± (0.0002) & \textbf{0.706} ± (0.0002) & \textbf{0.874} ± (0.0003) \\ 
 & GAIN & 0.9 ± (0.0004) & 1.098 ± (0.0003) & 1.075 ± (0.0002) & 1.108 ± (0.0001) \\ 
 & MICE & 1.731 ± (0.0073) & 1.74 ± (0.0086) & 1.781 ± (0.0046) & 1.303 ± (0.0112) \\ 
 & MissForest & 1.303 ± (0.0112) & 1.292 ± (0.0106) & 1.42 ± (0.0124) & 1.42 ± (0.0124) \\ 
LogNormal & Cond. Imp. GAN & \textbf{0.966} ± (0.0003) & \textbf{0.926} ± (0.0002) & \textbf{1.016} ± (0.0002) & \textbf{0.972} ± (0.0002) \\ 
 & GAIN & 0.997 ± (0.0003) & 0.958 ± (0.0003) & 1.139 ± (0.0001) & 1.242 ± (0.0001) \\ 
 & MICE & 13.468 ± (0.6489) & 13.856 ± (0.3277) & 14.022 ± (0.6454) & 9.187 ± (0.2773) \\ 
 & MissForest & 9.187 ± (0.2773) & 9.526 ± (0.1508) & 10.483 ± (0.0765) & 10.483 ± (0.0765) \\ 
Poisson & Cond. Imp. GAN & \textbf{0.74} ± (0.0003) & \textbf{0.781} ± (0.0001) & \textbf{0.794} ± (0.0002) & \textbf{0.852} ± (0.0002) \\ 
 & GAIN & 0.995 ± (0.0003) & 0.996 ± (0.0002) & 1.064 ± (0.0001) & 0.948 ± (0.0001) \\ 
 & MICE & 3.908 ± (0.0273) & 3.953 ± (0.017) & 4.039 ± (0.0127) & 2.979 ± (0.0371) \\ 
 & MissForest & 2.979 ± (0.0371) & 3.036 ± (0.0268) & 3.218 ± (0.0175) & 3.218 ± (0.0175) \\ 
Uniform & Cond. Imp. GAN & \textbf{0.756} ± (0.0003) & \textbf{0.808} ± (0.0002) & \textbf{0.937} ± (0.0003) & \textbf{0.865} ± (0.0002) \\ 
 & GAIN & 1 ± (0.0003) & 1.038 ± (0.0002) & 1.022 ± (0.0001) & 1.076 ± (0.0001) \\ 
 & MICE & 13.862 ± (0.1199) & 13.89 ± (0.0566) & 14.044 ± (0.0502) & 10.587 ± (0.0871) \\ 
 & MissForest & 10.587 ± (0.0871) & 10.512 ± (0.1881) & 11.13 ± (0.0836) & 11.13 ± (0.0836) \\ 
\hline \\[-1.8ex] 
\end{tabular}
}
\end{table} 

Conditional Imputation GAN yields the best RMSE overall and for each type of data distribution, performing fairly consistently for each feature. Our method performed particularly well for Gaussian, Poisson, and Uniform distributed features; the slightly higher RMSE for Log Normal and Exponential distribution is due to sampling from uniform initial starting values for the two right-skewed distributions, and is mitigated with longer training time. In practice, initial values can also be sampled from the distribution of observed values for each feature. In contrast, both of the non-GAN benchmarks MICE and MissForest have reasonable RMSE for Gaussian and Exponential distributions, but perform poorly with Log Normal and Uniform distributions. Furthermore, these two benchmarks had higher standard errors across distributions, an indication that GAN-based methods provide more robust imputations overall. These results validate the flexibility of Conditional GAN-based imputations given the ground truth of different underlying distribution by category. Conditioning on auxiliary information greatly improved imputation quality under non-MCAR mechanisms.

\section{MSR Ranking Data - Imputation Quality by Heterogeneous Subgroups}

\subsection{Data and Methodology}
\begin{figure}[h]
    \centering
    \includegraphics[width=0.8\textwidth]{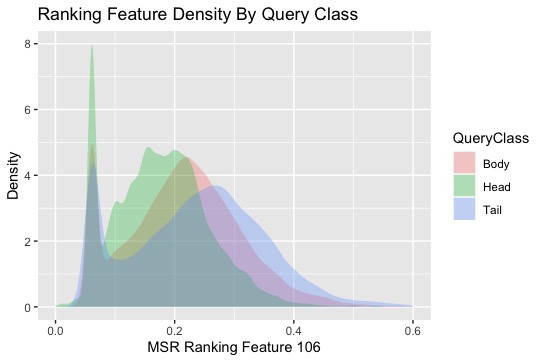}
    \caption{MSR - Ranking Feature 106 Distribution by Query Class. Note that tail queries display a lower concentrated peak between 0.05-0.1 and greater variance with more observations skewed to the right. Conditioning on Query Class allows for imputation with respect to these subgroups.}
    \label{fig:msr106_density}
\end{figure}

To demonstrate how the Conditional Imputation GAN accounts for heterogeneous subgroups within ranking data, we utilize the 10K query-group ranking dataset (MSLR-WEB10K) \citep{DBLP:journals/corr/QinL13} made available by Microsoft Research. Imputation quality is again measured by RMSE averaged across multiple imputations. The dataset is split into 80\% training and 20\% testing. Here note that each query-group can be categorized as Head, Body, or Tail queries depending on the query group size; this feature will be henceforth referred to as ``Query Class" and is also an always observed column in practice. Head and tail queries usually indicate very different subgroups and ranking feature distributions; see Table \ref{fig:msr106_density} for an example. Query Class also influences the probability of missingness. For example, in an e-commerce setting, tail queries indicate rare or newly-launched items with higher probability of missing values due to lack of data. This would violate the naive MCAR setting and is an example of the extended EMAR and EAMAR mechanisms.

Similar to the simulation experiment, we condition on Query Class and select four levels of missingness (5\%, 10\%, 20\%, 30\%), with values from 5 ranking features randomly masked as missing. In TensorFlow, we train each GAN replicate using the same hyperparameters and generator-discriminator architecture as the previous experiment. In R, we use default MissForest and MICE settings to implement the two missing data imputation methods. 10 imputations of test ranking dataset are generated for each method and missingness level.

%\begin{figure}[h]
    %\centering
    %\includegraphics[width=0.8\textwidth]{rmse_msr_ranking.jpeg}
   % \caption{MSR Results - Average RMSE by imputation level.}
   % \label{fig:rmse_msr}
%\end{figure}

\subsection{MSR Results}
Conditional Imputation GAN yields the lowest RMSE across the board, especially for higher missingness levels; it also learns ranking feature distributions with respect to auxiliary information such as Query Class. 
%See Figure \ref{fig:rmse_msr} and Table 
See Table \ref{table:rmse_msr_table}, which compares average RMSE and standard errors against benchmarks.

In terms of non-GAN benchmarks, MissForest outperformed MICE. However, we do not recommend implementing MissForest for a large ranking dataset in general from a computational standpoint. Given that it is based on random forests, implementation can be computationally expensive \citep{tang2017random}. On a standard Google Colab notebook, both GAN imputations took only about 1 minute to train for the full MSR ranking data, and similarly only about 1 minute to train for default settings of MICE (5 iteration per multiple imputation). Meanwhile, it took more than 10 minutes to run for default settings of MissForest (10 iterations) on just a 10\% MSR data sample, even though the data is low-dimensional. 
%\textcolor{red}{In terms of other GAN benchmarks, an additional experiment with MNIST images (not included) compared imputation quality between Conditional GAN and DC-GAN. The former demonstrated superior performance in terms of RMSE across various levels of missingness.}

\begin{table}[h] \centering 
  \caption{MSR Results - Imputation Quality (RMSE)} 
  \label{table:rmse_msr_table} 
\begin{tabular}{@{\extracolsep{0pt}} ccccc} 
\\[-1.8ex]\hline 
Method & 5\% Missing & 10\% Missing & 20\% Missing & 30\% Missing \\
\hline \\[-1.8ex] 
Cond. Imp. GAN & $\mathbf{0.232}$ & $\mathbf{0.237}$ & $\mathbf{0.239}$ & $\mathbf{0.239}$ \\ 
 & ($\pm 0.00001$) & ($\pm 0.00001$) & ($\pm 0.00001$) & ($\pm 0.00003$)\\ 
  \hline \\[-1.8ex]
GAIN & $0.234$ & $0.240$ & $0.240$ & $0.248$ \\ 
 & ($\pm 0.00002$) & ($\pm 0.0001$) & ($\pm 0.0001$) & ($\pm 0.0001$)\\ 
MICE  & $0.309$ & $0.311$ & $0.312$ & $0.314$ \\ 
 & ($\pm 0.0011$) & ($\pm 0.00057$) & ($\pm 0.00054$) & ($\pm 0.00044$) \\ 
MissForest & $0.240$ & $0.243$ & $0.249$ & $0.250$ \\ 
 & ($\pm 0.00227$) & ($\pm 0.0027$) & ($\pm 0.00277$) & ($\pm 0.00202$) \\ 
\hline \\[-1.8ex] 
\end{tabular} 
\end{table}

\section{Amazon Search Ranking Data}

%Since many real-world datasets suffer from incomplete data, we demonstrate the usefulness of imputed datasets generated by Conditional GANs in this ``learning-to-rank" setting. Our evaluation for this experiment will be two-fold: measuring imputation quality through comparing RMSE of imputed vs. true ranking data, and measuring ranking quality through comparing NDCG and MRR of ranking models trained on imputed vs. true ranking data.

%The Conditional GAN  will be used to generate ``fake" copies of a standard ranking dataset, analyze the performance (e.g., RMSE) across different levels of missingness, and compare the performance of standard ranking models (e.g., NDCG, MRR) trained on imputed vs.\ true ranking datasets. 
\subsection{Data and Methodology}
We now utilize an Amazon Search ranking dataset consisting of 1.8 million query-groups. This large ranking dataset is used to conduct experiments in order to answer two key questions: (1) \textit{What's the highest missingness level where Conditional Imputation GAN can still deliver good results?} (2) \textit{How does imputation quality translate to downstream applications, e.g., ranking quality?}

A collection of 24 common features for ranking are chosen and fall under 3 groups: behavioral, semantic, and product characteristics. Ranking features selected have varying ranges, units, and spread, which increases the difficulty of imputation. Each feature is normalized by the mean and standard deviation, maintaining the same correlation structure as the original features. Higher correlation is observed amongst ranking features that belong to the same group. For confidentiality reasons, individual feature names are masked and referenced as Feature A, B, C, etc. 

\begin{figure}[h]
  \centering
  \includegraphics[width=0.8\textwidth]{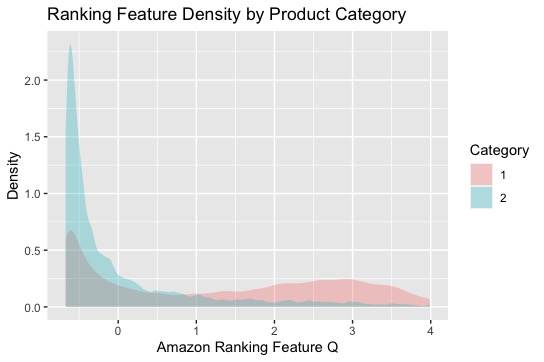}
\caption{Amazon - Ranking Feature Dist. by Category. Conditional distributions are significantly different, which motivates conditioning on product categories to improve imputation quality.}
\label{fig:histograms}
\end{figure}

Similar to previous experiments, the Amazon ranking dataset is also characterized by descriptive columns that are always observed and ranking features with potentially missing values. Descriptive columns includes a 21-class column of product categories, which are quite imbalanced; conditional on these observed values, ranking features can display significantly disparate underlying distributions. For example, a histogram of Feature Q across two product categories 1 and 2 is bi-modal and heavily right-skewed, respectively; see Figure \ref{fig:histograms} where conditional distributions are significantly different by both t-test and Kolmogorov-Smirnov, $p<0.05$. Hence, it is logical to condition on these product categories as auxiliary information to improve imputation quality.

\subsection{Imputation Quality: RMSE by Missingness Level}
To evaluate imputation quality by missingness level, we sample 500K query-groups for training and evaluate RMSE on a 300K query-group hold-out set. For each method, we specify product category columns that are always observed as auxiliary information to condition upon, and randomly mask ranking feature values as missing from 10\% to 90\%. This is aligned with the more expansive EMAR missing mechanism, of which MCAR is a special case.

10 imputations of the test set is generated for each missing percentage and method, and RMSE is computed based on imputed vs. true feature values. On a ml.m5.24xlarge AWS instance, each GAN replicate is trained for 1000 epochs using default Adam optimizer. The generator-discriminator architecture follows a standard CGAN architecture with fully connected layers, batch normalization, and leaky-relu activation. Imputation results generated by GAIN is included as a reference.

%The missing mechanism satisfies MCAR conditions, a special case of EMAR. Both GANs are trained for 1000 epochs using default Adam optimizer settings \citep{kingma2014adam}, with equivalent architecture consisting of 3 fully-connected layers, 2 dropout layers, mini-batch size = 512, leaky relu = 0.01, and dropout keep rate = 0.25. Computation time is 40 minutes to train a GAN on a ml.m5.24xlarge instance in AWS, which is $\approx$ 7 hours for one replicate with 9 missing percentage settings. 

%Multiple imputed copies of our golden ranking dataset are then generated for each missing percentage. 

\begin{figure}[h]
    \centering
    \includegraphics[width=0.8\textwidth]{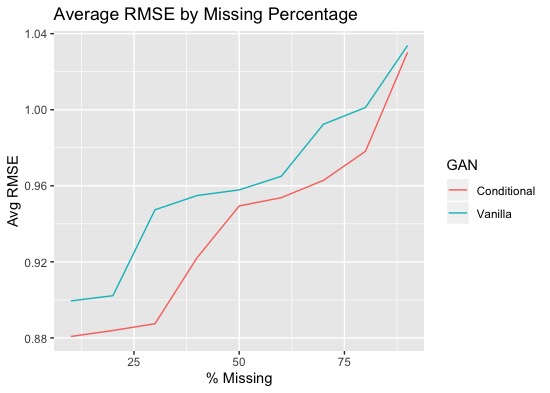}
    \caption{Amazon Results - Imputation Quality (RMSE) by Missingness Level. Conditional Imputation GAN imputations have lower errors than GAIN (Vanilla) on average at all missingness levels. }
    \label{fig:rmse_amazon}
\end{figure}

Figure \ref{fig:rmse_amazon} charts the increase in RMSE as missingness level is increased from 10\%; a good initial missing proportion would be 30\% or lower. The Conditional Imputation GAN results in lower RMSE across the board by better learning the disparate underlying distributions conditional on auxiliary columns, and is a testimony to the flexibility of GAN-generated imputations overall. In addition, our method also demonstrates comparable performance at higher missingness levels; the RMSE for 50\% missing is equivalent to that of GAIN at just 30\% missing.

%wouldn't list these numbers until specific section/experiment? 

\subsection{Ranking Quality: Ranking Models with Imputed Data}
To illustrate the downstream effect of imputed training data on ranking quality, we compare NDCG and MRR for standard ranking models trained on imputed versus baseline ground-truth models. We sample 500K query-groups each for training, testing, and validation sets. A standard model choice is the pairwise LambdaMart \citep{burges2010ranknet} model trained through LightGBM \citep{ke2017lightgbm}, a computationally efficient gradient boosting framework.

For the training set, we fix always observed columns and randomly mask 20\% of feature values as missing. 30 imputed copies of the ranking dataset are generated using the Conditional Imputation GAN with similar hyperparameters and generator-discriminator set-up as previous experiments; similar imputations are generated via GAIN as a reference point. Given the large-scale dataset and longer training time, we also optionally included dropout layers for regularization purposes. Other methods such as MissForest do not scale in this case due to computational cost; a 500K query-group dataset with 200+ results per group easily exceeds 100 million rows of data. 

The imputed ranking datasets are then used to train a standard ranking model using LightGBM, a total of 60 models with 30 for each GAN architecture. All models share the same binary target with the same testing and validation ranking set. Final ranking model metrics are directly comparable; specific computations of NDCG and MRR are discussed below.

\subsubsection{NDCG \& MRR} \label{sec:ndcg_mrr}
Ranking quality will be measured by two metrics, Normalized Discounted Cumulative Gain and Mean Reciprocal Rank. As a key information retrieval metric, Normalized Discounted Cumulative Gain (NDCG) \citep{valizadegan2009learning} measures ranking quality by summarizing the gains from a particular ranking order. It is standardized by position and is between $[0, 1]$. We first define Discounted Cumulative Gain (DCG) at position $p$, that is, for the top $p$ results returned by a ranking model as:
\begin{align*}
{\mathrm  {DCG_{{p}}}}=\sum _{{i=1}}^{{p}}{\frac  {2^{{rel_{{i}}}}-1}{\log_{{2}}(i+1)}}
\end{align*}
where $rel_{i}$ is the ranking score of result at position $i$ for query $q$, as predicted by an arbitrary ranking model. Greater penalty is given for relevant results ranked in lower positions. DCG is divided by Ideal Discounted Cumulative Gain (IDCG), the maximum possible DCG through position $p$. If ranking model orders a set of results in the optimal order possible, the NDCG will be equal to 1. That is,
\begin{align*}
    {\mathrm  {NDCG_{{p}}}}={\frac  {DCG_{{p}}}{IDCG_{{p}}}}, \ \ {\displaystyle \mathrm {IDCG_{p}} =\sum _{i=1}^{\vert rel_{p}\vert}{\frac {2^{rel_{i}}-1}{\log_{2}(i+1)}}}
\end{align*}
and $\vert rel_{p}\vert$ represents the optimal order of search results up to position $p$. This optimal order is usually known via the target column. The final NDCG given by a specific ranking model or search engine algorithm can be computed as the average of the NDCG for each query-group in the testing ranking dataset, and is directly comparable across different models or algorithms. 

The second metric, Mean Reciprocal Rank (MRR) \citep{radev2002evaluating}, measure ranking quality by evaluating the probability of the first correct answer for a given query. Specifically
\begin{align*}
    {\text{MRR}}={\frac  {1}{\vert Q \vert}}\sum _{{q=1}}^{{\vert Q \vert}}{\frac  {1}{{\text{rank}}_{q}}}
\end{align*}
where ${\text{rank}}_{i}$ is the rank position of the first relevant result for query $q$ in a dataset with $Q$ query-groups.

\subsubsection{Results}
From Table \ref{table: amazon_30rep_rmse}, imputations generated by a Conditional Imputation GAN with dropout layers resulted in the lowest RMSE on average and is 8.5\% lower ($p<0.05$) than the benchmark GAIN equivalent. The additional improvement from dropout illustrates that other neural net training techniques can further optimize imputation based on data type. 

\begin{table}[htp] \centering 
\caption{Amazon Results - Imputation Quality (RMSE), 30 Replicates} 
\label{table: amazon_30rep_rmse} 
\begin{tabular}{@{\extracolsep{0pt}} ccc} 
\\[-1.8ex]\hline 
\hline \\[-1.8ex] 
GAN Structure & Test RMSE & Std. Err.\\ 
\hline \\[-1.8ex]
Cond. Imp. GAN w/ Dropout & $\mathbf{0.881}$ & $0.00004$ \\ 
Cond. Imp. GAN & $0.930$ & $0.00004$  \\ 
GAIN w/ Dropout & $0.963$ & $0.00004$ \\ 
GAIN & $0.985$ & $0.00003$ \\ 
\hline \\[-1.8ex] 
\end{tabular} 
\end{table} 

We now compare ranking models trained on the imputed datasets, as measured by four performance metrics that indicate ranking quality. NDCG$_{10}$ and MRR$_{10}$ are defined in Section \ref{sec:ndcg_mrr} and computed based on the first 10 results per query-group as returned by the ranking model. NDCG and MRR Gain refer to percentage-wise improvement compared the baseline ranking model. In terms of all four metrics, the ranking models trained on imputations from Conditional GANs demonstrated statistically significant gains than those from the vanilla GAIN; see Table \ref{t-tests}. This result holds even after controlling for false discovery rates using Benjamini-Hochberg procedure. Unequal variances are assumed and Satterthwaite approximation for degrees of freedom (df*) is used. We emphasize here that given the volume of queries, even a minor but statistically significant improvement in performance is important for impact on downstream applications.

%Two-sample t-tests (Table \ref{t-tests}) showed the improvement in ranking quality by using Conditional GAN imputations was statistically significant, even after controlling for false discovery rates using Benjamini-Hochberg procedure. Unequal variances are assumed and Satterthwaite approximation for degrees of freedom (df*) is used. We emphasize here that given the volume of queries, even a minor but statistically significant improvement in performance is important for impact on downstream applications.

\begin{table}[htp] 
\centering 
  \caption{Amazon Results - Ranking Model Metrics: Conditional Imputation GAN vs. GAIN w/ Dropout}
  \label{table: amazon_ndcg_mrr} 
\begin{tabular}{@{\extracolsep{0pt}} ccccc} 
\\[-1.8ex]\hline 
\hline \\[-1.8ex] 
 & Mean Diff & t-stat & df* & $p$-val \\ 
\hline \\[-1.8ex] 
NDCG$_{10}$ & $0.001$ & $5.967$ & $39.769$ & $0$ \\ 
NDCG Gain & $0.172$ & $5.970$ & $39.800$ & $0$ \\ 
MRR$_{10}$ & $0.001$ & $5.336$ & $40.091$ & $0$ \\ 
MRR Gain & $0.185$ & $5.340$ & $40.100$ & $0$ \\ 
\hline \\[-1.8ex] 
\label{t-tests}
\end{tabular} 
%\footnotesize{*Unequal variance w/ Satterthwaite approx.}
\end{table}

\FloatBarrier

%Finally, we train a ground-truth ranking model based on the ``golden" 300K query group ranking dataset, using a binary target and 600 iterations to avoid underfitting. All hyperparameters for this ground-truth model is the same as that of the 60 models trained on imputed datasets. Given that we are learning on imputed datasets, we always expect to see a drop in NDCG compared to the ground-truth model. On average, the drop in NDCG for ranking models trained on imputations from Conditional GAN is $-2.41\% \pm 0.16\%$, while from vanilla GANs is $-2.71\% \pm 0.27\%$. These are fairly small differences in performance given the authors' prior expertise in training production ranking models. We provide the first known quantification of ranking quality differences due to missing data imputation, and conclude that ranking models trained on GAN imputed datasets in general are almost as good as training on the ground-truth dataset. 

%synthetic ranking experiment plots

\section{Conclusion}
We have demonstrated a novel Conditional Imputation GAN for extended missing mechanisms in ranking applications. Theoretical analysis showed compatible imputation guarantees for EMAR and EAMAR mechanisms that encompass a broader collection of missing models and datasets. Using a variety of ranking datasets, we showcase the superior imputation quality of our method against standard benchmarks. Experiment results illustrate the flexibility of the method, which generalizes well across a range of distributions and heterogeneous subgroups specified by always observed columns. GAN-based imputation approaches also scale computationally for very large datasets compared to the random-forest based MissForest, and generalizes better for complex missing mechanisms compared to the MCAR assumptions of MICE. 

In particular, simulations with five different distributions show that Conditional Imputation GAN outperformed traditional imputation methods such as MICE and MissForest, especially for non-Gaussian distributions. Furthermore, our method's imputations had lower standard errors overall which attests to the robustness across multiple imputations. Results using the open-source MSR ranking dataset confirm that Conditional Imputation GAN adapts well to multi-modal distributions that vary significantly conditional on auxiliary information, which other benchmarks fail to capture. Finally, experiments with the proprietary 1.8 million query-group Amazon ranking dataset demonstrate that downstream ranking models trained on imputed data also perform well as measured by NDCG and MRR. 

%Second, we demonstrate our methodology on a novel use case in e-commerce and search: learning-to-rank with incomplete data using a novel Amazon ranking dataset. Our Conditional GAN imputations have the lowest RMSE against benchmarks. With dropout layers, the RMSE of imputations from Conditional GANs is 8.5\% lower than from vanilla GANs. Training with GAN imputed data produce ranking models that are just as good as ground-truth ranking models based on standard ranking quality metrics NDCG and MRR. This illustrates a simple but versatile solution that could be broadly applicable to many downstream ML applications in the case of missingness in training data. 

%Finally, we replicate our methodology's success with the open-source MSR ranking dataset and a simulated ranking dataset with known distributional differences across product categories. For each experiment, Conditional GAN imputations produced the lowest overall RMSE across all levels of missingness compared to popular non-GAN benchmarks MICE and MissForest. Furthermore, our method performed consistently well across ranking features generated from five distributions (Gaussian, LogNormal, Exponential, Poisson, Uniform), and attests to the versatility of GAN-based imputations.

%Given that many real-world datasets consist of heterogeneous or non-standard ranking feature distributions and class labels, inclusion of this auxiliary information accounts for between-class variability and yields better imputation results.

Future work can explore whether GAN imputation optimality is possible under challenging missing mechanisms such as MNAR, and whether more complex GAN architecture (e.g. multi-task multi-label) could also benefit GAN imputation quality.

%A future experiment could include a formal A/B test to deploy ranking models trained on ground-truth vs. imputed data to production, and then formally test whether there are significant differences or impacts on business metrics of interest. 

\newpage

\nocite{*}
\bibliography{refs}

\end{document}